\documentclass{article}

\usepackage{arxiv}

\usepackage[ruled,vlined]{algorithm2e}
\usepackage{xcolor}

\usepackage[round,authoryear]{natbib}

\usepackage{xcolor}

\usepackage{hyperref}
\usepackage{url}

\usepackage{times}
\usepackage{helvet}
\usepackage{courier}
\usepackage{booktabs} 
\usepackage{makecell} 
\usepackage{tabularx} 

\usepackage{multirow} 
\usepackage{epsfig}
\usepackage{graphicx}
\usepackage{amsmath}
\usepackage{amssymb}

\usepackage[utf8]{inputenc} 
\usepackage[T1]{fontenc}    
\usepackage{hyperref}       
\usepackage{url}            

\usepackage{amsfonts}       
\usepackage{nicefrac}       
\usepackage{microtype}      
\usepackage{lipsum}		
\usepackage{natbib}
\usepackage{doi}

\title{SMOOT: Saliency Guided Mask Optimized Online Training}

\date{} 					

\author{ \href{https://orcid.org/0000-0000-0000-0000}{\includegraphics[scale=0.06]{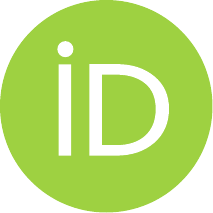}\hspace{1mm}Ali Karkehabadi} \\
	Department of Electrical and Computer Engineering\\
	University of California, Davis\\
	\texttt{akarkehabadi@ucdavis.edu} \\
	\And
	\href{https://orcid.org/0000-0000-0000-0000}{\includegraphics[scale=0.06]{orcid.pdf}\hspace{1mm}Houman Homayoun} \\
	Department of Electrical and Computer Engineering\\
	University of California, Davis\\
	\texttt{hhomayoun@ucdavis.edu} \\
        \And
	\href{https://orcid.org/0000-0000-0000-0000}{\includegraphics[scale=0.06]{orcid.pdf}\hspace{1mm}Avesta Sasan} \\
	Department of Electrical and Computer Engineering\\
	University of California, Davis\\
	\texttt{asasan@ucdavis.edu} \\
}

\renewcommand{\headeright}{}
\renewcommand{\undertitle}{}
\renewcommand{\shorttitle}{}

\begin{document}
\maketitle

\begin{abstract}
Deep Neural Networks are powerful tools for understanding complex patterns and making decisions. However, their black-box nature impedes a complete understanding of their inner workings. Saliency-Guided Training (SGT) methods try to highlight the prominent features in the model's training based on the output to alleviate this problem. These methods use back-propagation and modified gradients to guide the model toward the most relevant features while keeping the impact on the prediction accuracy negligible. SGT makes the model's final result more interpretable by masking input partially. In this way, considering the model's output, we can infer how each segment of the input affects the output. In the particular case of image as the input, masking is applied to the input pixels. However, the masking strategy and number of pixels which we mask, are considered as a hyperparameter. Appropriate setting of masking strategy can directly affect the model's training. In this paper, we focus on this issue and present our contribution. We propose a novel method to determine the optimal number of masked images based on input, accuracy, and model loss during the training. The strategy prevents information loss which leads to better accuracy values. Also, by integrating the model's performance in the strategy formula, we show that our model represents the salient features more meaningful. Our experimental results demonstrate a substantial improvement in both model accuracy and the prominence of saliency, thereby affirming the effectiveness of our proposed solution.
\end{abstract}

\section{Introduction}
The transformative influence of deep learning on our lives stems from its ability to learn from data and discover complex patterns in complex datasets. Deep Neural Networks (DNN) have enabled us to make more accurate predictions and make better decisions supported by data, serving as a catalyst for societal and technological evolution. Despite their unquestionable utility, the black-box nature of DNNs raises concerns about the trustworthiness and reliability of their outputs and the facets affecting their inference function. Consequently, there exists a significant interest to understand the behavior of these models and identify specific features they use and prioritize when producing an outcome." Generating a reliable explanation is especially important in sensitive domains such as medicine (\cite{caruana2015intelligible}), neuroscience, finance, and autonomous driving (\cite{li2018tell}). Not only do these explanations contribute to the critical understanding and trustworthiness of models, but they also aid in the process of model debugging (\cite{zaidan2008modeling,li2018tell}) and model tuning. In light of the aforementioned, a substantial volume of scholarly research has been dedicated to the development of interpretability methods aimed at comprehending the internal mechanisms of DNNs (\cite{bach2015pixel,kindermans2016investigating,smilkov2017smoothgrad,singla2019understanding,singh2017hide,lecun2015deep}). A prevailing approach to interpreting model decisions involves identifying significant input features that highly influence the final classification decision (\cite{baehrens2010explain,singla2019understanding,smilkov2017smoothgrad,zeiler2014visualizing,selvaraju2017grad}). Commonly referred to as saliency maps, these methods typically employ gradient calculations to assign an importance score to individual features, thus reflecting their impacts on the model's prediction (\cite{selvaraju2017grad,shrikumar2017learning}). 
 
Saliency maps can be unclear due to noise or distracting elements, which makes them harder to comprehend and less accurate. To address this issue, (\cite{singh2017hide}) proposed explanation methods that leverage higher-order backward gradients to give insight into the saliency maps. Also, (\cite{kindermans2016investigating}) benefits from multiple gradient calculations. An example is the SmoothGrad technique, which mitigates saliency noise by repetitively adding noise to the input and subsequently averaging the resulting saliency maps for each input (\cite{singla2019understanding}). Other techniques like integrated gradients(\cite{smilkov2017smoothgrad}), DeepLIFT(\cite{selvaraju2017grad}), and Layer-wise Relevance Propagation(\cite{bach2015pixel}) modify the backpropagation through a different gradient function (\cite{ancona2017towards}).  However, these methods' effectiveness is intrinsically tied to their reliability and stability (\cite{adebayo2018sanity,kindermans2016investigating}). If saliency maps change drastically for slight perturbations in the input or model, their trustworthiness can be severely compromised (\cite{ghorbani2019interpretation}). Thus, in developing novel interpretability techniques, it is imperative to establish robust and comprehensive sanity checks to ensure their validity and (\cite{adebayo2018sanity,thorne2018fever}). Furthermore, the quality of explanations generated by these methods can vary significantly depending on the data type (images, text, time series, etc.) and the model architecture (CNN, Recurrent Neural Networks, Transformer-based models, etc.). Hence, it's crucial to adapt and develop new interpretation techniques considering these factors (\cite{ismail2020benchmarking,sundararajan2017axiomatic}). Moreover, the quest for better interpretability extends beyond understanding individual predictions. It's about deciphering the learned representations and decision-making logic of the model as a whole (\cite{hooker2019benchmark,ross2017right}). Neural network distillation into interpretable models like soft decision trees has been studied as a means to improve interoperability(\cite{frosst2017distilling}).

\section{Related Works} 
Interpretability in machine learning refers to the ability to understand a learning model's decisions and actions and explain them in human-comprehensible terms(\cite{chakraborty2017interpretability}). This research paper focuses on the advancement and extension of existing gradient-based methods that serve to define the behavior of models. Such methods have demonstrated their value in facilitating knowledge transfer and enabling the post-hoc interpretation of models.
Building upon prior works in this domain, we aim to enhance the effectiveness and applicability of gradient-based approaches in understanding and interpreting model behavior while improving the model's generalization by selecting the most robust features during model training. In this section, we review the relevant prior-art literature on interpretability and saliency-guided training. 

\subsection{interpretability} 
Previous research has explored different approaches to enhance the performance and interpretability of neural networks. Perkins et al. proposed a grafting method for feature selection, by incrementally expanding the feature set while training a prediction model using gradient descent. This technique accelerates the regularized learning process, making it suitable for large-scale applications (\cite{perkins2003grafting}). Frosst et al. introduced a method for yielding a soft decision tree through stochastic gradient descent, leveraging neural network predictions (\cite{frosst2017distilling}). Young et al. created a benchmark dataset with annotated "rationales" provided by humans, allowing for the measurement of model justifications against human justifications (\cite{deyoung2019eraser}). In (\cite{ghaeini2019saliency}), Ghaeini et al. proposed saliency learning aiming to train models that make predictions with explanations that align with that of the ground truth. Ross et al. developed a method to effectively explain and regularize differentiable models by penalizing input gradients based on expert annotations in an unsupervised manner (\cite{ross2017right}). Wang et al. introduced a solution that emphasizes class discrimination as a fundamental component in training CNNs for image classification, improving model discriminability and reducing visual confusion (\cite{wang2019sharpen}). (\cite{devries2017improved}) demonstrated the cutout regularization technique to enhance CNN's robustness and performance. Behrens et al. proposed a method to explain local decisions made by arbitrary classification algorithms, utilizing local gradients as estimated explanations (\cite{baehrens2010explain}).

\subsection{Saliency Guided Training}
In the saliency-guided training, Ismail et al. (\cite{ismail2021improving}) introduce a new algorithm incorporating interpretability to enhance models' accuracy and saliency. They established their algorithm based on employing gradients for saliency map detection. Gradients are representative of the degree of "importance" attributed to the input pixels.

Although gradient methods have a lot of positive characteristics when used with visual models, they frequently result in noisy pixel attributions in areas unrelated to the predicted class (\cite{kapishnikov2021guided}). Potentially meaningless local fluctuations in partial derivatives may be the cause of the noise observed in a sensitivity map (\cite{smilkov2017smoothgrad}). The gradient of the model, when trained using a common method based on empirical risk minimization (ERM), may change drastically in response to minor input perturbations. Low-level features contain object position information but are intermingled with noises like backdrops, whereas high-level features display rich semantic information but lack object position information (\cite{liu2022improved}). In order to accomplish saliency map extraction using low-level features like color and texture., downstream algorithms need more precise criteria. This is where the saliency guidance training approach comes in. Saliency-guided training aims to decrease the gradient values and maintain the model performance. This approach reduces the impact of irrelevant features on the outcome of the model as the gradients get closer to zero. By incorporating saliency guidance into the training procedure, the model can focus on the most significant aspects of the data, potentially leading to improved performance and generalization. 

Algorithm \ref{SGT_Original} describes the SGT process which uses saliency information in training a neural network model $f_{\theta}$. In this algorithm $\mathcal{D_{KL}}(p \| q)$ is the KL divergence between probability distributions $p$ and $q$. the $\mathcal{D_{KL}}$ quantifies the difference between the original output distribution $f_{\theta}(X)$ and the modified output distribution $f_{\theta}(\widetilde{X})$. The $M_k(I, X)$ is the masking function that removes the bottom $k$ features from the input data $X$, based on the sorted index $I$ representing the importance of features according to their gradients. The $\widetilde{X}$ is the input data with the least important $k$ features masked out. It is obtained by applying the masking operation $M_k(I, X)$. The $L_i$ is the combined loss function used for training. It includes two terms: the standard loss term $\mathcal{L}(f_{\theta}(X), y)$ that measures the model's performance on the original input $X$ with corresponding labels $y$, and a regularization term involving the KL divergence to encourage similarity between the output distributions of $X$ and $\widetilde{X}$.

\begin{algorithm}
    \SetAlgoNlRelativeSize{0}
    \caption{{Saliency Guided Training (Original)}}
    Require Training samples $X$, number of features to be masked $k$, learning rate $\tau$, hyperparameter $\lambda$\\
    Initialize $f_{\theta}$ \\
    Preload or randomize for new training\\
    \For{$i = 1$ \textbf{to} epochs}{
        \For{$i = 1$ \textbf{to} epochs}{
            \textcolor{blue}{Calculate the sorted index $I$ for the gradient of output w.r.t the input.} \\
            $I = \text{sort}(\nabla_X f_{\theta_i}(X))$ \\
        
            \textcolor{blue}{Mask the bottom $k$ features of the original input.} \\
            $\widetilde{X} = M_k(I, X)$\\ 
            
            \textcolor{blue}{Compute the loss function with regularization term.} \\
            $L_i = \mathcal{L}(f_{\theta_i}(X), y) + \lambda \mathcal{D_{KL}}  (f_{\theta_i}(X) \| f_{\theta_i}(\widetilde{X}))$
            
            \textcolor{blue}{Update network parameters using the gradient.}\\
            $f_{\theta_{i+1}} = f_{\theta_i} - \tau \nabla_{\theta_i} L_i$ 
        }
    }
\label{SGT_Original}
\end{algorithm}

    
        
        

\subsection{Motivation and Problem Statement}

In Algorithm \ref{SGT_Original}, the number of masked features is determined using parameter $k$. Authors in \cite{ismail2021improving}, although noted that parameter k may affect the SGT optimization, have assumed this value constant. The solution in this paper was motivated after we investigated the impact of $k$ and realized that the best choice of $k$ is input image dependent. 

\begin{figure}[hbt!]
\vspace{-7pt}
  \centering
  \includegraphics[width=0.7\columnwidth]{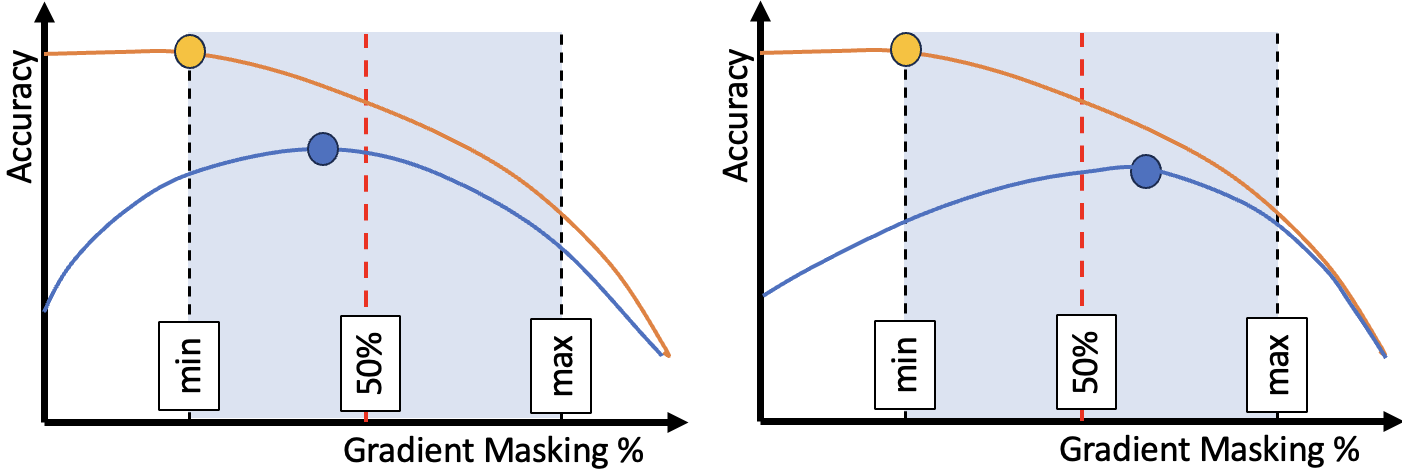}
  \vspace{-8 pt}
  \caption {Illustration of how sorted gradient masking could result in a monotonic decrease in accuracy in majority of images (in orange), but an initial increase and then decrease in accuracy in some other images (in blue). The figure on the left captures the case where the peak accuracy in the exception images is reached before the 50\% masking point, and the figure on the right captures the case where peak accuracy in the exception images is reached after 50\% masking point.}
 \label{param-def1}
\end{figure}

In our motivational experiment leading to the solution proposed in this paper, we first used backpropagation to compute the gradient value for all input pixels. We then sorted the pixels based on the gradient value.  We then started masking pixels with the highest gradient. The highest gradient pixels are expected to be the most important input features affecting the decision of the learning model. 
We expected to see a monotonically reducing accuracy curve as we masked additional input features. Interestingly, we observed a different behavior for some input images, in which the accuracy of the model started increasing by masking the high-value gradients initially, and then after reaching a max value, it started reducing.
This behavior is captured in both images shown in Figure \ref{param-def1}. The orange curve represents the behavior of the majority of images, where by increasing the making percentage, the classification accuracy monotonically reduces, and the blue curve represents the exception minority where the peak accuracy is reached after some initial masking. Figure \ref{param-def1} also illustrates that the peak in accuracy in the exception group can happen before (left) or after (right) the 50\% making point.  

The next important question we attempted to investigate was the frequency at which such a scenario transpires. To answer this question we trained a two-layer CNN model on the CIFAR-10 dataset featuring a kernel size of 3 and a stride of 1, succeeded by two fully connected layers. For regularization, we introduced two dropout layers with rates set at 0.25 and 0.5. Our experiment indicated that a considerable portion of the images, 16\% of test images in our specific experiment, fall into this category. This indicates that the observed behavior is not statistically negligible.  This observation also holds logical validity. The dimensions and orientation of an object within an image impact the quantity of input pixels integrated into the input feature, thereby triggering activation of the input image. This indicates that the optimal value for $k$ is contingent upon the specific image. Additionally, when examining the saliency map of an input image, it is frequently noted that there are prominent pixels within the saliency map that lie outside or deviate from the area of interest within the image. This suggests that the model is allocating attention to features that hold no significant relevance for its classification objective. This observation likely elucidates why the removal of certain pixels initially leads to an increase in model accuracy, until the authentic features are discerned, ultimately resulting in a decline in accuracy. Stemming from this exploratory experimentation, we postulated that the parameter $k$ in Algorithm \ref{SGT_Original} assumes a pivotal role as a hyperparameter, enabling the model to learn better features, while also enhancing the alignment between the saliency map and the location of the object of interest. Given this motivational background this paper addresses the following problem statement:\\[-8pt]

\textbf{Problem Statement}:
Given a learning model $f_{\theta}$ and data $\{(X_i,y_i)\}_{i=1}^n$, formulate a saliency-guided training solution that optimizes the number of masked input pixels ($k$) based on saliency metrics and adjust model parameters $\theta$ to enhance the model accuracy and saliency map's fidelity.

\vspace{-8 pt}
\section{Methodology}
To enhance the model's generalization and the accuracy of the saliency map by focusing on a broader, more crucial set of features, as well as to minimize noise by teaching the model to ignore irrelevant features for input classification, we introduce an upgraded Saliency Guided Mask Optimized Online Training (SMOOT) approach. In this approach, the hyperparameter $k$ specifies the number of masked pixels.

In the context of the Algorithm \ref{SGT_Original}, the authors fixed parameter K. Their proposal advocates for configuring this parameter to encompass 50\% of the pixels. To illustrate, when applied to images of dimensions 28*28, this parameter is set at 392 pixels. As explained in the motivational and problem statement section of this paper, for some images, gradient masking initially increases the classification accuracy and then reduces it. As mentioned, and illustrated in Figure \ref{param-def1}, the max accuracy could also be realized by dropping less (left image), or more (right images) than 50\% of high gradient pixels. Given this observation, we propose starting with 50\% gradient masking and modifying the Algorithm \ref{SGT_Original} to dynamically adjust the number of masked pixels in the direction where accuracy is maximized. 

Our proposed solution, SMOOT, is outlined in Algorithm \ref{SGT_updated}. In this approach, instead of a single hyperparameter $k$, we use a vector $K_i$, where $K_i(X)$ is the masking percentage of input image $X$ in epoch i. The objective is to optimize model parameters by finding the best masking percentage for each input image to minimize the loss function.  The algorithm starts by initializing model parameters $f_{\theta}$ and by assigning each value in $K$ vector to 50\%. We then adjust the masking percentage for each image in the direction that image accuracy increases. As discussed previously, there are two types of images. images where increasing the masking percentage monotonically decreases their classification accuracy. Let us denote this group of images as class I images. We also have the images where masking initially increases the accuracy and then results in a drop in accuracy. Let us denote these images as class II images. 

For class I images, if we adjust the masking in the direction where accuracy increases, we always move towards the direction of reducing the masking percentage. We define a fixed threshold for minimum (min) and maximum (max) masking. In this case, for the class, I group, the masking percentage always shifts towards the min masking value, as as shown in Figure \ref{param-def1} stays there. For class II, however, the move is towards the direction of max accuracy. As shown in Figure \ref{param-def1}, the move could be to the left or the right until the accuracy max (shown with blue dot) in Figure \ref{param-def1} is achieved.

\begin{algorithm}
    \SetAlgoNlRelativeSize{0}
    \caption{{SMOOT: Saliency Guided Mask Optimized Online Training}}
    Training samples $X$, learning rate $\tau$, hyperparameter $\lambda$, hyperparameter $\alpha$, controls increase or decrease number of masking $\mu$\\
    Initialize $f_{\theta}$: Preload or randomize for new training\\
    Initialize $K$: 50\% to be consistent with prior work\\
    \For{$i = 1$ \textbf{to} epochs}{
        \For{$i = 1$ \textbf{to} epochs}{
    
            \textcolor{blue}{Get sorted index $I$ for the gradient of output w.r.t the input.} \\
    
            \textbf{1.} $I = \text{sort}(\nabla_X f_{\theta_i}(X))$ \\    
            
            \textcolor{blue}{Compute $\widetilde{X}$ as the image with bottom $k$ features of the original input masked.}

            \textbf{2.} $\widetilde{X} = M(i, K, I, X)$  \\
            \textcolor{blue}{Compute difference in softmax outputs when $softmax_i$ is ith highest softmax output} 
    
            \textbf{3.} $\delta_1 = softmax_1(\widetilde{X}) - softmax_1(X)$  \\
            \textbf{4.} $\delta_2  = \frac{1}{n-1}\sum_{i=2}^{i=n}(softmax_i(\widetilde{X}) - softmax_i(X))$ \\
            \textbf{5.} $\delta = \alpha\delta_1 + (1-\alpha)\delta_2$ \\    
                
            \textcolor{blue}{Find number of masking for next epoch}\\
            \textbf{6.} $K_{i+1}(X) = \max(K_{\min}, \min(K_{\max}, K_{i} + \lfloor \mu\delta \rfloor))$ \\
            \textcolor{blue}{Compute the loss function} \\
            \textbf{7.} $L_i = \mathcal{L}(f_{\theta_i}(X), y) + \lambda \mathcal{D_{KL}}  (f_{\theta_i}(X) \| f_{\theta_i}(\widetilde{X}))$ \\
            \textcolor{blue}{Use the gradient to update network parameters} \\
            \textbf{8.} $f_{\theta_{i+1}} = f_{\theta_i} - \tau \nabla_{\theta_i} L_i$ 
        }
    }
\label{SGT_updated}
\end{algorithm}

        
            
            

Let us define $softmax_i(X)$ to be the softmax output of image X in epoch i. Let's also define $\widetilde{X}$ to be the masked image $X$. In this case, the difference between the accuracy of top 1, top 5, or a combination of the 2 could be used to indicate if masking results in improvement or degradation in accuracy.  The change in the top 1 accuracy is computed in line 5 of the SMOOT algorithm as:

\begin{equation}
\delta_1 = softmax_1(\widetilde{X}) - softmax_1(X)
\end{equation}
and the change in the top 2 to top $n$ is computed using  
\begin{equation}
\delta_2  = \frac{1}{n-1}\sum_{i=2}^{i=n}(softmax_i(\widetilde{X}) - softmax_i(X))
\end{equation}
In this equation, for top 5 accuracy, n should be equal to 5. We then use a weighted representation of change in softmax value using the equation:   
\begin{equation}
\delta = \alpha\delta_1 + (1-\alpha)\delta_2
\end{equation}
For the generated results, in the result section of this paper, we have used the $n=5$ and $\alpha=0.7$, placing more priority on improvement in top 1 accuracy. Given the change in the accuracy, we then change the masking percentage of the input image using the following equation: 
\begin{equation}
  K_{i+1}(X) = \max(K_{\min}, \min(K_{\max}, K_{i} + \lfloor \mu\delta \rfloor)) 
\end{equation}
In this equation, $K_{min}$ and $K_{max}$ are the min and max percentages allowed for masking. In our experiment $K_{min}=20$ and $K_{max}=80$. Finally, the weight "$\mu$" is a hyperparameter that determines the speed at which the masking percentage changes. The loss of the model is computed similarly to the previous SGT using:
\begin{equation}
    L_i = \mathcal{L}(f_{\theta_i}(X), y) + \lambda \mathcal{D_{KL}} (f_{\theta_i}(X) \| f_{\theta_i}(\widetilde{X}))
\end{equation}

In which $\mathcal{L}$ is the cross entropy loss and $\mathcal{D_{KL}}$ is the KL divergence. The KL divergence or relative entropy is computed using:
\begin{equation}
\small
    \mathcal{D_{KL}}(f_{\theta_i}(X)\|f_{\theta_i}(\widetilde{X}))=\sum_{x\in X}f_{\theta_i}(X)log(\frac{f_{\theta_i}(\widetilde{X})}{f_{\theta_i}(X)})
\end{equation}

The KL divergence is computed based on the similarity of $f_{\theta_i}(X)$ to $f_{\theta_i}(\widetilde{X})$, first term representing the output for the original input image $X$, and second representing the output for masked image $\widetilde{X}$ using the $K_i(X)$ masked in the current epoch. Using this updated loss, the gradients are then used to update the network parameters as follows:
\begin{equation}
            f_{\theta_{i+1}} = f_{\theta_i} - \tau \nabla_{\theta_i} L_i
\end{equation}

The algorithm of our proposed solution, including all steps above is shown in Algorithm \ref{SGT_updated}. The algorithm takes as input the training samples ($X$), the initial number of features to be masked ($k$), learning rate ($\tau$), and a hyperparameter ($\lambda$). It starts by initializing the model parameters ($f_{\theta}$). Then, for each epoch and mini-batch, the following steps are executed. 

First, in line 1, the sorted index $I$ corresponding to the gradient of the output concerning the input, denoted as $\nabla_X f_{\theta_i}(X)$, is found. In this algorithm, $sort(\cdot)$ is a sorting function. $sort(\nabla_X f_{\theta}(X))$ denotes the sorted gradient. In line 2, we generate a new masked image from the input image. $M(\cdot)$ represents the input masking function. $M(i,K,I,X)$ generate $X(i)$ with a mask distribution, removing the $K(i)$ lowest features as indicated by sorting vector I to generate $\widetilde{X}$. In lines 3-6, based on the contrast between the accuracy of the input and the masked input, the algorithm determines and applies the magnitude and direction of change in the masking percentage.  In line 7 the loss function is computed by adding a weighted KL divergence to the standard loss term. The KL divergence is computed by comparing the output of the model when using original input X versus masked input Xm, using the mask value $K_i(X)$ generated in epoch i. The weighted KL divergence terms account for The saliency-guided regularization in the overall loss function in line 7. Finally, the network parameters are updated using gradient descent in line 8.
\section{Experiments and Results}
Our primary objective is to assess the performance of SMOOT relative to SGT and traditional models across different datasets. To this end, we evaluate the efficacy of our proposed solution by retraining models on the MNIST (\cite{lecun2010mnist}), Fashion MNIST (\cite{xiao2017fashion}), CIFAR-10, and CIFAR-100 (\cite{krizhevsky2009learning}) datasets.

\subsection{Model Architecture}
For the MNIST and Fashion-MNIST Datasets, we employed a two-layer CNN with a kernel size of 3 and a stride of 1. The CNN layers were succeeded by two fully connected layers. Additionally, two dropout layers were incorporated with rates of 0.25 and 0.5. For the CIFAR datasets, we turned to the Tiny Transformer, adopting the original configurations: a minimalist 'tiny' dimension of $(L=12, d=192, h=3)$. The pre-trained model for this setup was sourced from the Facebook research repository\footnote{\url{https://github.com/facebookresearch/deit}} and is based on the \textit{deit} architecture, which was pre-trained on ImageNet (\cite{he2016deep}). As the concluding layer, we incorporated a 10-neuron classifier. Both architectures were trained on a single NVIDIA A100 GPU for a span of 100 epochs with batches of 256. Adadelta (\cite{zeiler2012adadelta}) served as our optimizer, operating at a learning rate of 1 for MNIST and Fashion-MNIST datasets and 0.001 for the CIFAR datasets. Throughout the training process, for each epoch, low gradient value pixels were substituted by random values within the spectrum of the remaining pixels. These gradients were computed using the Captum library (\cite{kokhlikyan2020captum}).\\
In the context of our paper, with regard to the MNIST and Fashion MNIST datasets, which inherently exhibit lower complexity, it is appropriate to configure the hyperparameter ($\alpha$) to a high value, such as 0.95. This strategic decision enables us to place a greater emphasis on the label. For the CIFAR-10 and CIFAR-100 datasets, we recommend setting the hyperparameter ($\alpha$) to a substantial value, like 0.8. This choice effectively guides the model to allocate more attention to classes beyond the label. Because the SGT article (\cite{ismail2021improving}) considered $K = 50\%$ of the total pixel count, in our paper, we also adopted an initial value of n equal to 50\% of the total pixels. Specifically, the initial value of n was set to 392 for the MNIST dataset and 512 for CIFAR. We set the $\mu$ to 10 and $\lambda$ to 1, following the approach described in SGT's paper (\cite{ismail2021improving}).
Table \ref{table:hyper} displays the model architecture and hyperparameters utilized in our paper, which are based on the SGT.

\begin{table} [hbt!]
  \centering
  \scalebox{0.8}{
  \begin{tabularx}{\linewidth}{l|X|X|X|X|X}
    \toprule
    \bf Dataset  & \bf Model & \bf Initialize $K$  & \bf $\tau$ & \bf $\alpha$ & \bf $\lambda$ \\
    \midrule
    MNIST &CNN & 392 & 1 &  95\% & 1\\
    Fashion-MNIST & CNN  & 392 & 1 &95\% & 1 \\
    CIFAR10 & Transformer & 512 & 0.001 &80\% & 1 \\
    CIFAR100 & Transformer & 512 & 0.001 &80\% & 1 \\
    \bottomrule
  \end{tabularx}
  }
  \caption{Model Architecture}
  \label{table:hyper}
\end{table}

\subsection{Saliency Guided Training for Images}
\begin{figure}[h]
    \centering
    \includegraphics[width=0.95\columnwidth]{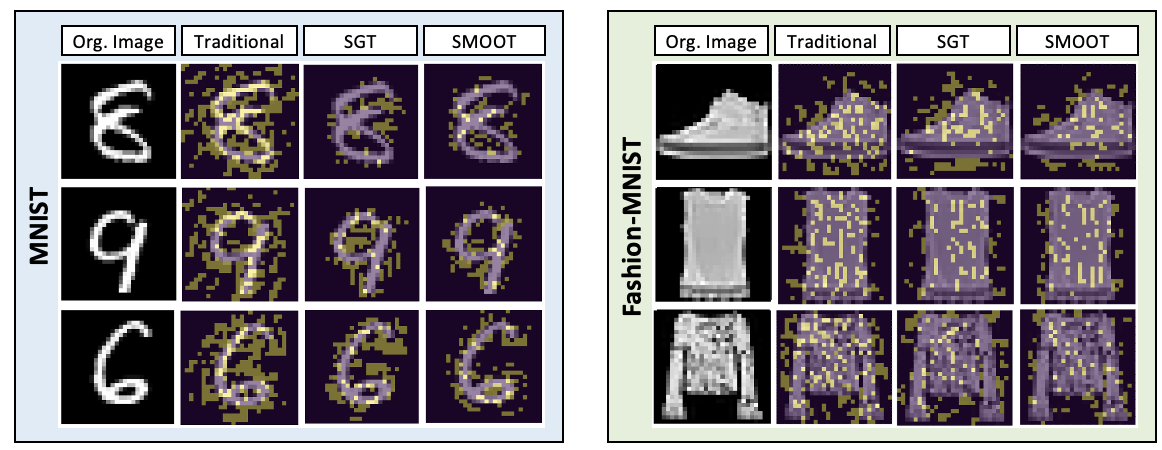}
    \vspace{-8 pt}
    \caption{MNIST and Fashion-MNIST: The displayed images compare our model (SMOOT) with an SGT and Traditional model. For saliency, the best model employs a high gradient to focus on the most prominent pixels. In the case of MNIST and Fashion- MNIST, it prioritizes edge detection.}
    \label{fig:CMNIST}
\end{figure}

In the context of image classification using saliency, it is common to encounter redundant features that are not crucial for the model's prediction. Take the example of an object's background in an image, which occupies a significant portion but typically holds little relevance to the classification task. When the model's attention is directed toward the object itself, it is desirable for the background gradient (representing most of the features) to be close to zero, indicating its diminished importance. Figure \ref{fig:CMNIST} illustrates a comparison between the saliency map generated using our approach, the SGT in (\cite{ismail2021improving}), and Traditional training (no saliency-guided training). Figure \ref{fig:CMNIST} provides this comparison for images selected from the MNIST dataset and Fashion MNIST dataset.  
    
\subsection{Model Accuracy Drop}
In our study, we investigate the efficacy of our approach, SMOOT, against the prior art SGT (\cite{ismail2021improving}) and conventional baseline training methods. We utilize various saliency techniques for this purpose, including modification-based evaluation (\cite{petsiuk2018rise,kindermans2016investigating}). We evaluated the impact on model accuracy by ranking and eliminating features based on their saliency values. The experiment is conducted on datasets such as MNIST and Fashion MNIST, both of which have known uninformative feature distributions (e.g., black background). The results show that SMOOT leads to a steeper drop in accuracy compared to traditional training and SGT, regardless of the saliency method employed. This steeper drop indicates that our method is more effective in identifying and eliminating less informative features, resulting in a more refined model. However, it's important to emphasize that this experiment is most relevant for datasets with known uninformative features and might not be applicable to datasets with varying or unknown backgrounds. Because in datasets with varying or unknown backgrounds, the uninformative features may not be as consistent or easily distinguishable. This means that saliency methods might struggle to effectively rank and eliminate them, potentially leading to different results.

\begin{table}[hbt!]
  \centering
  \scalebox{0.8}{
  \begin{tabularx}{\linewidth}{l|X|XXX|XX}
    \toprule
    \bf MNIST  & \bf \# Test & \bf Min(K) & \bf Median(K) & \bf Max(K) & \bf Accuracy & \bf AUC \\
    \midrule
    Traditional & 10K & 0 & 0 & 0  & 99.40\% &  36.35 \\
    SGT & 10K  & 392 & 392 & 392 & 99.35\% &  34.67 \\
    SMOOT & 10K & 234 & 388 & 544& 99.40\% &  33.16 \\
    \bottomrule
  \end{tabularx}
  }
  \caption{MNIST: Traditional and Saliency Model Training Accuracy and saliency (A smaller AUC value indicates better performance in saliency)}
  \label{tabel:T_MNIST}
\end{table}

Table \ref{tabel:T_MNIST} provides an overview of the Area Under accuracy drop Curve (AUC) and accuracy results on the MNIST dataset for different gradient-based training approaches. The table compares traditional training, training with a saliency-guided procedure, and our own method. 
It's noteworthy that a lower AUC value is indicative of superior performance, largely due to its representation of a more pronounced drop in accuracy. Upon careful analysis of the table, it is clear that SMOOT demonstrates superior accuracy and saliency when compared to the traditional and SGT approach.

Table \ref{tabel:T_FashionMNIST} displays the outcomes of AUC and the achieved accuracy on the Fashion-MNIST dataset. Analyzing the table, it becomes clear that our model exhibits superior accuracy and saliency in comparison to both the SGT and traditional models, as indicated by the smaller AUC values.

\begin{table} [hbt!]
  \centering
  \scalebox{0.8}{
  \begin{tabularx}{\linewidth}{l|X|XXX|XX}
    \toprule
    \bf Fashion-MNIST  & \bf \# Test & \bf Min(K) & \bf Median(K) & \bf Max(K) & \bf Accuracy & \bf AUC \\
    \midrule

    Traditional &10K & 0 & 0 &  0 & 93.60\% & 40.79 \\
    SGT & 10K  & 392 & 392 & 392 &93.35\% & 39.91 \\
    SMOOT & 10K & 223 & 372 & 576 &93.65\% & 36.18 \\
    \bottomrule
  \end{tabularx}
  }
  \caption{Fashion-MNIST: Comparing Traditional and Saliency and SMOOT Model Training Accuracy and Saliency (Smaller AUC = Better Saliency)}
  \label{tabel:T_FashionMNIST}
\end{table}

\begin{figure}[hbt!]

  \centering
  \includegraphics[width=0.8\columnwidth]{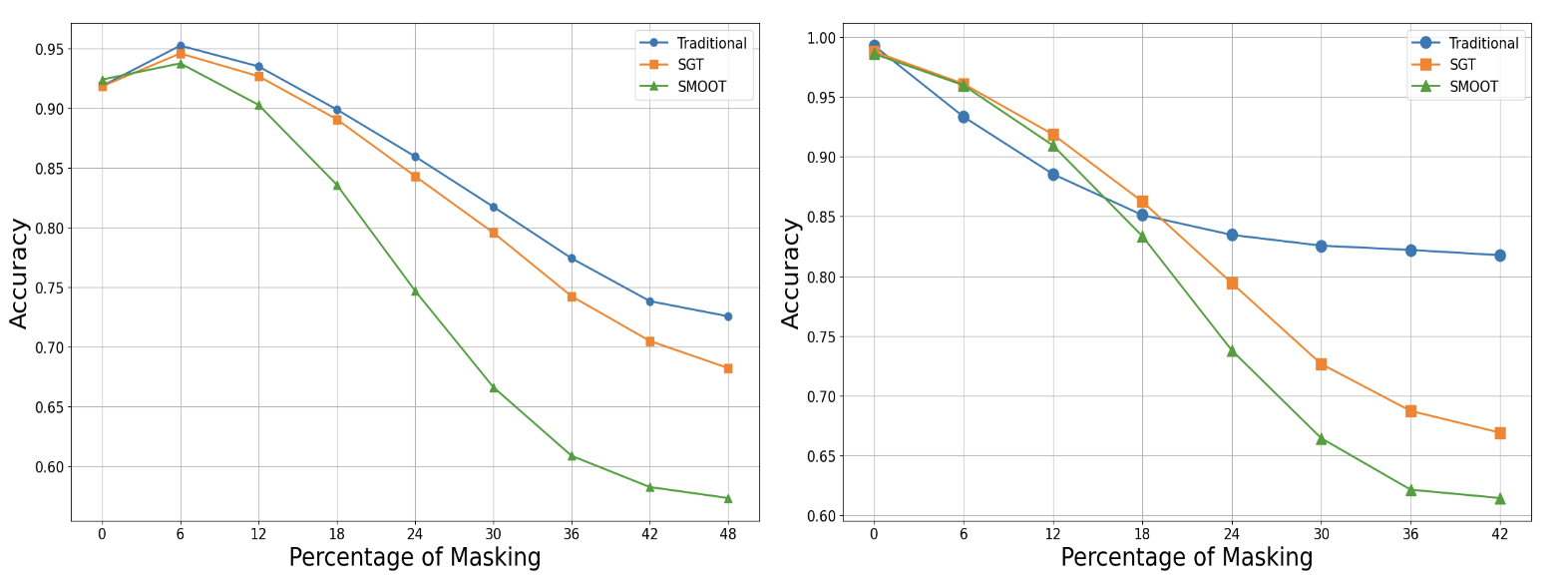}  
  \vspace{-8 pt}
  \caption {Comparing accuracy drop For MNIST and Fashion-MNIST using three training approaches: Standard cross-entropy based training, SGT (\cite{ismail2021improving}), and Our proposed (SMOOT). A more pronounced decrease is indicative of superior performance.}
 \label{MNIST}
\end{figure}

Based on the changes in accuracy due to the number of masking, we present our findings in Figure \ref{MNIST} for MNIST and Fashion-MNIST datasets. These figures illustrate that our model exhibits a more significant drop in accuracy compared to both the traditional and SGT models, demonstrating its higher Salient in comparison to them.

\subsection{Saliency Guided Mask Optimized Online Training for Transformers}

\begin{figure}[hbt!]
    \centering
    \includegraphics[width=0.97\columnwidth]{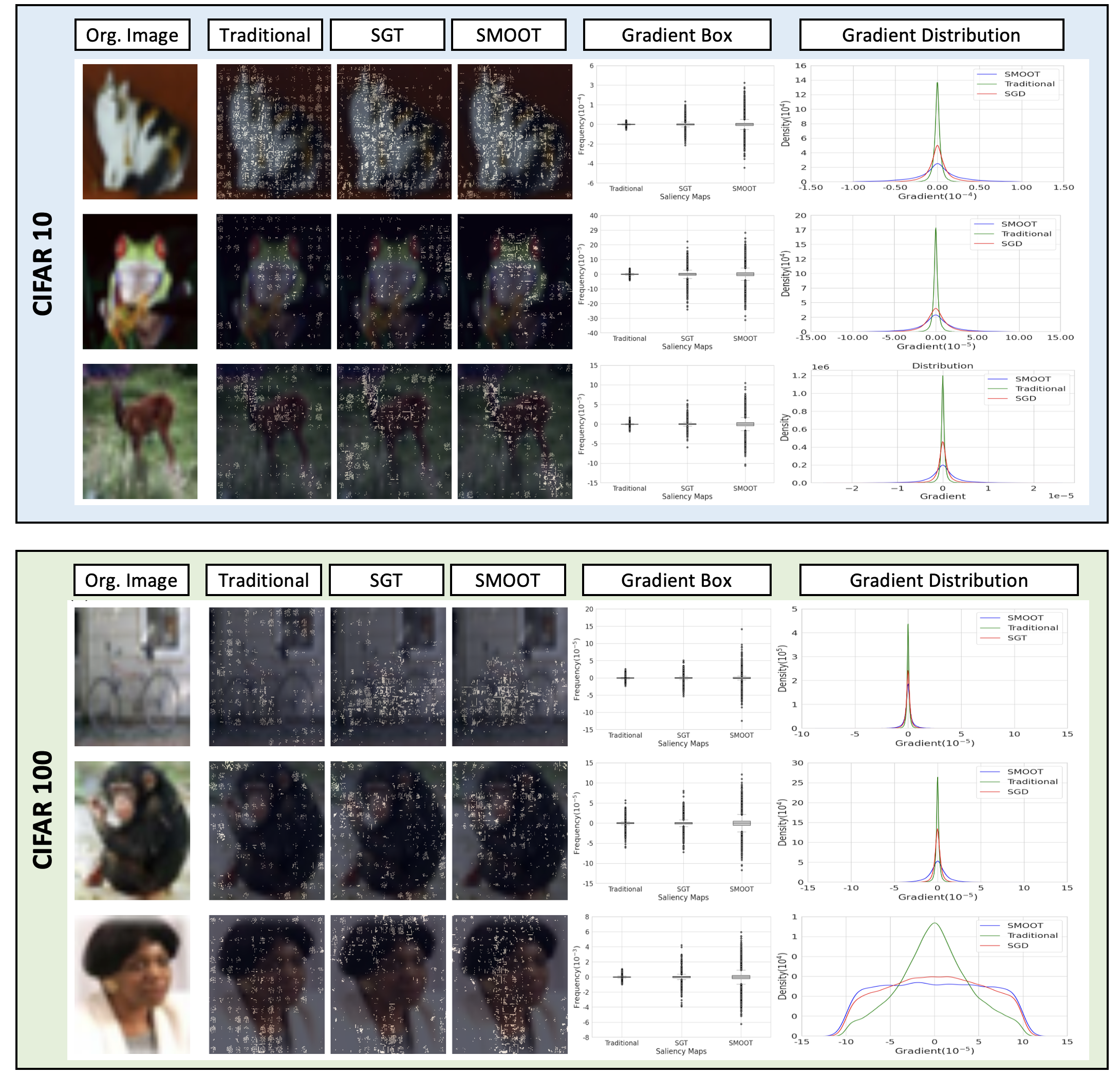}
    \caption{CIFAR10 and CIFAR100: The displayed image compares our model (SMOOT) with a  Saliency-Guided Training (SGT) and Traditional model by using a transformer. For saliency, the best model employs a high gradient to focus on the most prominent pixels. Gradients approaching zero often signify uninformative features, whereas exceptionally large or exceedingly small gradient values tend to highlight the informativeness of these features.}
    \label{fig:transformer}
\end{figure}

The Transformer model represents a revolutionary computational paradigm in the realm of deep learning, demonstrating robust performance across an array of computer vision tasks. With its roots in natural language processing (NLP), the Transformer model(\cite{vaswani2017attention}) has shown an unparalleled ability to capture long-range dependencies through the self-attention mechanism. Its related large-scale counterparts, such as BERT (\cite{devlin2018bert}), GPT-3 (\cite{tomsett2020sanity}) and GPT-4 (\cite{openai2023gpt4}), have set new standards in harnessing powerful language representations from unlabeled textual data. This remarkable success of Transformer models in the NLP landscape has piqued the curiosity of the vision community, leading to its effective application in various computer vision tasks. These encompass areas like image recognition (\cite{dosovitskiy2020image,touvron2021going}), object detection (\cite{carion2020end}), and even image generation (\cite{chen2021pre}). Within this transformative framework, the Tiny Transformer (\cite{touvron2022deit})., stands out as a streamlined variant. This model is distinguished by its significantly reduced size, all while maintaining performance levels comparable to its standard Transformer counterparts. In Table \ref{tabel:transformer_cifar10} and \ref{tabel:transformer_cifar100}, the implementation of the SGT method resulted in CIFAR10 and Cifar100 in a notable improvement in accuracy. Notably, through the application of the SMOOTH method and optimization of masked values, we were able to achieve even higher accuracy values, Furthermore, Figure \ref{fig:transformer} presents a comparison among traditional, SGT, and SMOOT models through the application of saliency maps to images.

\begin{table} [hbt!]
  \centering
    \scalebox{0.8}{
  \begin{tabularx}{\linewidth}{l|X|XXX|X}
    \toprule
    \bf CIFAR10  & \bf \# Test & \bf Min(K) & \bf Median(K) & \bf Max(K) & \bf Accuracy  \\
    \midrule
    Traditional & 10K  & 0 & 0 & 0 & 95.65\% \\
    SGT & 10K  & 512 & 512 & 512 & 96.05\% \\
    SMOOT & 10K & 204 & 488 & 753 & 96.35\%  \\
    \bottomrule
  \end{tabularx}
  }
  \caption{CIFAR10: The accuracy compares our model (SMOOT) with a  Saliency-Guided Training (SGT) and Traditional model by using a transformer. }
  \label{tabel:transformer_cifar10}
\end{table}

\begin{table} [hbt!]
  \centering
  \scalebox{0.8}{
  \begin{tabularx}{\linewidth}{l|X|XXX|X}
    \toprule
    \bf CIFAR100  & \bf \# Test & \bf Min(K) & \bf Median(K) & \bf Max(K) & \bf Accuracy  \\
    \midrule
    Traditional & 10K  & 0 & 0 & 0 & 75.75\% \\
    SGT & 10K  & 512 & 512 & 512 & 78.10\% \\
    SMOOT & 10K & 362 & 432 & 682 & 79.65\%  \\
    \bottomrule
  \end{tabularx}
  }
  \caption{CIFAR100: The accuracy compares our model(SMOOT) with a  Saliency-Guided Training(SGT) and Traditional model by using transformer. }
  \label{tabel:transformer_cifar100}
\end{table}

\section{Conclusion}
In the framework of SGT, the hyperparameter $k$ is of paramount importance as it represents the "number of masking." This hyperparameter is instrumental in the learning process by pinpointing the most relevant pixels in each image. Also, such identification proves vital for improving accuracy, especially in datasets with larger image dimensions. Therefore, careful optimization of $k$ is crucial to ascertaining its optimal value. In this paper, we present a novel solution for input-based selection of $k$, illustrating that such optimization improves both the accuracy of the model and the fidelity of the saliency map across all tested benchmarks.

\bibliographystyle{unsrtnat}
\bibliography{template}  

\end{document}